# Physics-Informed GMDH for Adaptive Field Reconstruction

**Mykhailo Minin**

JSC KIEP, Kyiv, Ukraine

Corresponding author: mykhailo.minin@gmail.com

## Abstract

Physics-informed computational methods usually optimize parameters within a functional representation whose structure is fixed in advance. This work proposes a Physics-Informed Method of Group Data Handling (PI-GMDH), in which representations of coupled physical fields are progressively constructed during solution. Candidate functional directions are evaluated through the first variation of the complete physical and observational objective, introduced in packages, and followed by block-coordinate damped Gauss-Newton coefficient optimization. The framework is demonstrated with tensor-product Chebyshev functions on the incompressible Navier-Stokes equations using a two-dimensional time-dependent Taylor-Green benchmark. Under the tested configuration, adaptive PI-GMDH reached validation and held-out test losses of 5.299e-19 and 5.296e-19 with 204, 201, and 175 active functions for u, v, and p. Complete degree-by-degree and all-terms PI-GMDH variants, together with selected PINN and KAN reference configurations, are used to examine the effect of structural construction policy. The Taylor-Green experiments demonstrate that package-wise structural-growth policies can substantially affect the attained reconstruction accuracy, representation size, and computational cost under the tested configuration. These results support PI-GMDH as a self-organizing basis-construction framework for physics-informed regression and field reconstruction; they do not separately establish the advantage of its candidate-scoring mechanism or general superiority over other physics-informed approaches.

## 1. Introduction

Physics-informed computational methods incorporate known physical relations directly into the construction or optimization of an approximation. Physics-informed neural networks (PINNs), for example, represent unknown fields by neural networks and include residuals of governing differential equations together with available observations, boundary conditions, and initial conditions in the optimization objective [3,4]. Related physics-informed learning methods have broadened this principle beyond a single neural architecture.

A common feature of many such approaches is that the general form of the approximation is chosen before optimization begins. Training then determines parameters within that representation. This raises a separate structural question: rather than only asking which parameter values minimize the physical objective, can the governing equations also determine which functional components should constitute the approximation?

The present work addresses this question through a Physics-Informed Method of Group Data Handling (PI-GMDH). Classical GMDH introduced self-organizing structural construction in which candidate models are generated, parametrically fitted, evaluated, and selectively retained as model complexity develops [1,2]. PI-GMDH adopts this self-organizing principle at the level of functional representations of physical fields: the governing physical equations participate directly in deciding which functional components should become active.

### 1.1 Relation to Existing Approaches

**GMDH and differential models.** GMDH and its descendants have previously been used for polynomial self-organization and for data-driven differential modeling. In particular, differential polynomial neural networks extend GMDH-type structures and have been used to construct or approximate partial-differential relations from observed data [12,13]. The present work therefore does not claim that differential equations have not previously appeared in GMDH-derived methods. The distinction is that the governing PDE is assumed known and its complete residual is used to construct the functional representation of the unknown solution fields.

**Residual-driven basis enrichment.** Adaptive finite-element, multiscale, and reduced-order methods also use residual or error information to enrich approximation spaces. Residual-based online enrichment in generalized multiscale finite-element methods, for example, constructs additional basis functions from the current PDE residual and problem data [14]. PI-GMDH shares the objective of adapting the approximation space, but its elementary candidates are evaluated directly as functional perturbation directions through the first variation of the complete coupled physics-informed objective, rather than being generated from local residual subproblems or solution snapshots.

**Greedy approximation in function space.** Stagewise additive modeling and gradient boosting interpret approximation as optimization in function space and progressively introduce components associated with descent directions [15]. This viewpoint is mathematically close to the present structural-selection mechanism. In PI-GMDH, however, a candidate perturbation is propagated through the governing differential operators and all coupled residual components before its relevance is evaluated. Candidate selection is therefore based on its first-order influence on the complete physical and observational objective.

**Physics-informed spectral representations.** Chebyshev and other orthogonal-polynomial representations are increasingly used in physics-informed neural, KAN, and operator-learning formulations [9,16-18], while adaptive spectral methods can modify active spectral content during solution [19]. Consequently, neither Chebyshev approximation nor physics-informed spectral representation is claimed as the novelty of the present work. Chebyshev functions are used here as a differentiable candidate family through which the proposed self-organizing structural mechanism can be studied.

The central PI-GMDH cycle consists of two coupled operations. First, the functional representation is structurally extended by introducing a package of new functional components. Second, the coefficients of the enlarged representation are reoptimized against the complete physical and observational objective. The package-construction rule is therefore an explicit component of the method rather than an incidental implementation detail. Different rules produce different PI-GMDH variants while retaining the same underlying physical formulation and parameter-optimization machinery.

A package may contain every function in the next level of a predefined functional hierarchy, a single function, the entire remaining candidate space, or a selectively chosen subset. The present work develops a physics-guided adaptive package rule in which candidate functions are assessed through the first variation of the complete objective. The governing equations are consequently used not only to optimize coefficients after functions have been activated, but also to decide which functional directions are useful for subsequent structural growth.

The proposed framework is not restricted to the Navier-Stokes equations. It requires a differentiable residual describing the physical problem and candidate functional directions for which the required operator responses can be evaluated. The incompressible Navier-Stokes equations are selected here because they provide a demanding coupled nonlinear test containing multiple fields, nonlinear convection, incompressibility, pressure-velocity coupling, and several derivative orders. A two-dimensional time-dependent Taylor-Green flow is used as a controlled benchmark with an analytical solution.

The principal contributions of this work are:

- Formulation of PI-GMDH as a general framework in which physics-informed coefficient optimization is coupled to progressive construction of the functional representation.
- Explicit formulation of structural growth through functional packages, allowing alternative package-construction policies to be investigated within a common PI-GMDH framework.
- Development of a physics-informed structural-selection mechanism in which candidate functional components are evaluated through the first variation of the complete coupled physical and observational objective, and selected components are introduced as packages before block-coordinate coefficient reoptimization.
- Integration of structural construction with block-coordinate coefficient optimization, allowing coupled physical fields to develop different active functional representations while remaining linked through the same physical objective.
- A controlled Navier-Stokes benchmark that compares PI-GMDH package-construction strategies using a common Chebyshev family and optimization mechanism, with selected PINN and KAN configurations included as external physics-informed references.

The individual ingredients have precedents in GMDH, residual-driven basis enrichment, greedy function-space approximation, and physics-informed spectral methods. This work formulates PI-GMDH as a self-organizing framework for constructing functional representations through physics-informed coefficient fitting and package-wise structural growth. The proposed implementation evaluates candidate directions through the first variation of the complete coupled objective. The experimental contribution is an examination of how alternative package-construction policies affect accuracy, representation size, and computational cost on a controlled Taylor-Green reconstruction problem. These comparisons assess the complete construction-and-optimization procedures; they do not independently establish the benefit of each component of their integration.

The numerical study therefore addresses a more specific question than whether one functional family is preferable to another: given a common candidate family and a common physics-informed objective, how does the policy used to construct packages of functional components affect the evolution, computational cost, and attainable accuracy of the representation?

## 2. Physics-Informed Method of Group Data Handling

### 2.1 Physical problem and complete objective

Consider independent variables $\boldsymbol{\xi} = (\boldsymbol{\xi}_1,\dots,\boldsymbol{\xi}_d)$ and $K$ unknown physical fields

$$\mathbf{f}(\boldsymbol{\xi}) = [f_1(\boldsymbol{\xi}),\dots,f_K(\boldsymbol{\xi})]^T \tag{1}$$

The physical problem is assumed to be expressible as

$$\mathcal{H}(\mathbf{f},\mathcal{D}\mathbf{f};\boldsymbol{\xi}) = 0 \tag{2}$$

where H may contain differential equations, algebraic constraints, conservation relations, boundary conditions, initial conditions, or other known physical relations, and Df denotes the required operators acting on the fields.

Let $\hat{f} = [\hat{f}_1,\dots,\hat{f}_K]^T$ denote the current approximation. Substitution into the governing relations defines a physical residual $r^{phys}$. When observations are available, their mismatch with the corresponding approximated quantities defines a data residual $r^{data}$. These contributions are assembled into the complete residual

$$\mathbf{r}(\hat{\mathbf{f}}) = \left[\mathbf{r}^{\text{data}};\mathbf{r}^{\text{phys}}\right] \tag{3}$$

A weighted least-squares objective is

$$E(\hat{\mathbf{f}}) = \mathbf{r}(\hat{\mathbf{f}})^T W \mathbf{r}(\hat{\mathbf{f}}), \quad W = W^T \succ 0 \tag{4}$$

Conventional parameter optimization reduces $E$ within a representation selected in advance. PI-GMDH additionally permits the functional representation itself to evolve.

### 2.2 Functional representation and packages

For each physical field, let the current approximation be represented by an active set $A_k$,

$$\hat{f}_k(\boldsymbol{\xi}) = \textstyle\sum_{i\in A_k} a_{k,i}\,\phi_i(\boldsymbol{\xi}) \tag{5}$$

where $\{\phi_i\}$ is an available or progressively explored functional family. The active sets of different physical fields need not be identical.

A structural extension of field k introduces a package

$$P_k \subseteq \{\phi_i : i \notin A_k\} \tag{6}$$

after which $A_k \leftarrow A_k \cup P_k$ and the coefficients of the enlarged representation are reoptimized. PI-GMDH therefore follows the generic cycle

current representation → candidate exploration → package construction →
structural extension → coefficient optimization (7)

The definition of $P_k$ is deliberately not fixed by the general framework. A package may be constructed from a predefined hierarchy, from one candidate at a time, from the complete remaining candidate space, or from a physics-guided selection criterion. This separation permits the influence of structural policy to be studied independently of the functional family and coefficient optimizer.

## 2.3 First variation and candidate residual response

For physics-guided package construction, consider an elementary candidate function $\phi_i$ for field k. Its local effect is examined through the infinitesimal perturbation

$$\hat{f}_k \to \hat{f}_k + \epsilon\phi_i, \qquad \epsilon \to 0 \tag{8}$$

The first-order response of the complete residual is

$$\mathbf{j}_{k,i} = \frac{d}{d\epsilon}\mathbf{r}\left(\hat{f}_1, \dots, \hat{f}_k + \epsilon\phi_i, \dots, \hat{f}_K\right)\Big|_{\epsilon=0} \tag{9}$$

Thus $\mathbf{j}_{k,i}$ is the Gâteaux derivative of the complete residual with respect to perturbation of field k in the functional direction $\phi_i$. Although the candidate is introduced into a single field, its residual-response vector contains its first-order influence on every coupled residual component.

The relevance of the candidate follows directly from the first variation of the complete objective:

$$\frac{dE}{d\epsilon}\Big|_{\epsilon=0} = 2\mathbf{j}_{k,i}^T W\mathbf{r} \tag{10}$$

The same physics-informed objective used to estimate active coefficients can therefore be used to determine whether an inactive functional direction is locally relevant to the remaining residual.

## 2.4 Adaptive package construction

Candidate directions are compared using the normalized residual alignment

$$q_{k,i} = \frac{|\mathbf{j}_{k,i}^T W\mathbf{r}|}{\sqrt{(\mathbf{j}_{k,i}^T W\mathbf{j}_{k,i})(\mathbf{r}^T W\mathbf{r})}} \tag{11}$$

The normalization removes dependence on the magnitudes of the current residual and candidate-induced residual response, so $0 \leq q_{k,i} \leq 1$ measures their absolute weighted directional alignment.

A second quantity retains information about the absolute interaction magnitude,

$$m_{k,i} = \frac{|\mathbf{j}_{k,i}^T W\mathbf{r}|}{N} \tag{12}$$

where $N$ is the number of sampled points used in the corresponding objective evaluation. Since $\mathbf{j}_{k,i}{}^{\mathrm{T}}\mathrm{W}\mathrm{r}$ accumulates weighted contributions of all residual components over those points, $m_{k,i}$ expresses this interaction on a per-sampled-point basis while retaining the weighted combination of residual components.

The two measures play complementary roles. $q$ identifies directions aligned with the current residual independently of scale, while $m$ measures the mean magnitude of their first-order interaction. Both are state dependent: the usefulness of a candidate changes as the coupled field approximations and residuals evolve.

An adaptive package can therefore be constructed by screening candidates according to $q$, prioritizing eligible candidates according to $m$, and introducing a bounded group before reoptimization. The threshold, package-size rule, and hierarchy-exploration rule are implementation parameters rather than defining properties of PI-GMDH itself.

## 2.5 Coefficient optimization

After structural extension, the coefficients of the active representation are reestimated. For field k, let

$$J_k = \left[\mathbf{j}_{k,1}\ \mathbf{j}_{k,2}\ \cdots\ \mathbf{j}_{k,n_k}\right] \tag{13}$$

contain residual-response columns associated with its active functions. With the remaining physical fields held fixed,

$$\mathbf{r}(\mathbf{a}_k + \delta\mathbf{a}_k) \approx \mathbf{r} + J_k\delta\mathbf{a}_k \tag{14}$$

The weighted linearized least-squares correction is

$$\delta\mathbf{a}_k = \arg\min_{\delta\mathbf{a}}(\mathbf{r} + J_k\delta\mathbf{a})^T W(\mathbf{r} + J_k\delta\mathbf{a}) \tag{15}$$

A damped Gauss-Newton step satisfies

$$(J_k^T W J_k + \lambda I)\delta\mathbf{a}_k = -J_k^T W\mathbf{r}, \qquad \mathbf{a}_k \leftarrow \mathbf{a}_k + \delta\mathbf{a}_k \tag{16}$$

Updating one field while holding the remaining field representations fixed follows the block-coordinate optimization principle. Block-coordinate optimization itself is not introduced as a new method here; it is the parameter-optimization mechanism coupled to structural construction. [11]

Structural and parameter optimization are linked by

$$(J_k^T W\mathbf{r})_i = \mathbf{j}_{k,i}^T W\mathbf{r} \tag{17}$$

Before activation this quantity determines the first variation associated with a candidate functional direction. After activation the same residual-response vector becomes a Jacobian column used to estimate coefficient corrections.

### 2.6 Package-construction policies

Four package policies are useful for the present study. In adaptive physics-guided construction, $P_k$ contains a selected subset identified from $q$ and $m$. In complete degree-by-degree construction, $P_k$ contains every function in the next degree shell. In one-term construction, $P_k$ contains one functional component. In all-terms construction, the complete remaining candidate family through a prescribed maximum complexity is introduced in one package.

These variants are treated as controlled PI-GMDH realizations or ablations. They are not claimed to be classical GMDH algorithms. Their purpose is to isolate the consequences of the package-construction policy within a common physics-informed structural framework.

## 3. Application to the Incompressible Navier-Stokes Equations

### 3.1 Governing equations and residuals

Consider an incompressible velocity field $u$(x,t) = [$u_1$,…,u_d]$^T$ and pressure $p$(x,t). For constant kinematic viscosity $\nu$,

$$\nabla \cdot \mathbf{u} = 0 \tag{18}$$

$$\frac{\partial \mathbf{u}}{\partial t} + (\mathbf{u}\cdot\nabla)\mathbf{u} - \nu\nabla^2\mathbf{u} + \nabla p = \mathbf{g} \tag{19}$$

For current approximations $\hat{u}$ and $\hat{p}$,

$$r_{\text{div}} = \sum_{b=1}^{d} \frac{\partial \hat{u}_b}{\partial x_b} \tag{20}$$

$$r_{m_a} = \frac{\partial \hat{u}_a}{\partial t} + \sum_{b=1}^{d} \hat{u}_b \frac{\partial \hat{u}_a}{\partial x_b} - \nu \sum_{b=1}^{d} \frac{\partial^2 \hat{u}_a}{\partial x_b^2} + \frac{\partial \hat{p}}{\partial x_a} - g_a \tag{21}$$

When observations of velocity component $u_a$ are available, $r_{u_a} = \hat{u}_a - u_a^{\text{obs}}$. Other observational, boundary, or initial-condition residuals can be included in the same manner.

### 3.2 Residual response to a velocity candidate

For a candidate $\phi_i$ considered for velocity component $\hat{u}$_a, the observation and incompressibility responses are

$$\left.\frac{dr_{u_a}}{d\epsilon}\right|_0 = \phi_i, \qquad \left.\frac{dr_{\text{div}}}{d\epsilon}\right|_0 = \frac{\partial \phi_i}{\partial x_a} \tag{22}$$

The momentum residual associated with the perturbed component responds as

$$\left.\frac{dr_{m_a}}{d\epsilon}\right|_0 = (\phi_i)_t + \sum_b \hat{u}_b\,(\phi_i)_{x_b} + \hat{u}_{a,x_a}\phi_i - \nu\nabla^2\phi_i \tag{23}$$

For c ≠ a, nonlinear convection also gives

$$\left.\frac{dr_{m_c}}{d\epsilon}\right|_0 = \hat{u}_{c,x_a}\phi_i \tag{24}$$

A candidate introduced into one velocity field therefore generally affects several blocks of the coupled residual.

### 3.3 Residual response to a pressure candidate

For $\hat{p} \rightarrow \hat{p} + \epsilon\phi_i$, incompressibility is unchanged, whereas

$$\left.\frac{dr_{m_a}}{d\epsilon}\right|_0 = \frac{\partial \phi_i}{\partial x_a} \tag{25}$$

In the absence of direct pressure observations or another pressure constraint, the physical residual determines pressure only through its spatial derivatives. The transformation $\hat{p}$(x,t) → $\hat{p}$(x,t)+g(t) leaves $\nabla\hat{p}$ and therefore the momentum residuals unchanged.

### 3.4 Two-dimensional specialization

$$r_{\mathrm{div}} = \hat{u}_x + \hat{v}_y \tag{26}$$

$$r_x = \hat{u}_t + \hat{u}\hat{u}_x + \hat{v}\hat{u}_y - \nu\left(\hat{u}_{xx} + \hat{u}_{yy}\right) + \hat{p}_x - g_x \tag{27}$$

$$r_y = \hat{v}_t + \hat{u}\hat{v}_x + \hat{v}\hat{v}_y - \nu\left(\hat{v}_{xx} + \hat{v}_{yy}\right) + \hat{p}_y - g_y \tag{28}$$

If both velocity components are observed, $r_u$ = $\hat{u} - u$^obs and $r_v$ = $\hat{v} - v$^obs, and one convenient residual ordering is r = [$r_u$,$r_v$,$r_{div}$,$r_x$,$r_y$]$^{\mathrm{T}}$.

For a candidate $\phi_i$ considered for $\hat{u}$, $\hat{v}$, and $\hat{p}$ respectively, the complete residual responses are

$$\mathbf{j}_{u,i} = \left[\phi_i, 0, (\phi_i)_x, (\phi_i)_t + \hat{u}_x\phi_i + \hat{u}(\phi_i)_x + \hat{v}(\phi_i)_y - \nu\Delta\phi_i, \hat{v}_x\phi_i\right]^T \tag{29}$$

$$\mathbf{j}_{v,i} = \left[0, \phi_i, (\phi_i)_y, \hat{u}_y\phi_i, (\phi_i)_t + \hat{u}(\phi_i)_x + \hat{v}_y\phi_i + \hat{v}(\phi_i)_y - \nu\Delta\phi_i\right]^T \tag{30}$$

$$\mathbf{j}_{p,i} = \left[0,0,0, (\phi_i)_x, (\phi_i)_y\right]^T, \qquad \Delta\phi_i = (\phi_i)_{xx} + (\phi_i)_{yy} \tag{31}$$

These expressions make explicit that candidate relevance is determined by the complete coupled system and changes as the current field approximations evolve.

## 4. Taylor-Green Benchmark and Numerical Implementation

### 4.1 Analytical benchmark

$$u^*(x, y, t) = \sin x \cos y \exp(-2\nu t) \tag{32}$$

$$v^*(x, y, t) = -\cos x \sin y \exp(-2\nu t) \tag{33}$$

$$p^*(x, y, t) = \frac{1}{4}[\cos(2x) + \cos(2y)]\exp(-4\nu t), \qquad \nu = 0.01 \tag{34}$$

$$p_x^* = -\frac{1}{2}\sin(2x)\exp(-4\nu t), \qquad p_y^* = -\frac{1}{2}\sin(2y)\exp(-4\nu t) \tag{35}$$

The benchmark uses the full spatial cell x,y ∈ [−π,π] and the temporal interval from 0 to 1. Before Chebyshev evaluation, each spatial coordinate is divided by $\pi$ and time is affinely mapped from [0,1] to [−1,1].

### 4.2 Training, validation, and test protocol

The revised experimental protocol contains 250,000 training points, 50,000 validation points, and 100,000 held-out test points, sampled independently using random seeds 43, 44, and 45, respectively. Model checkpoints are selected using validation loss, and the selected checkpoint is subsequently evaluated on the held-out test set.

Exact velocity values $u*$ and $v*$ provide the observational information entering the physics-informed objective. Pressure and pressure gradients are not supplied as direct training targets. Pressure reconstruction is therefore assessed through its coupling to the velocity fields in the momentum equations.

$$L = \frac{1}{N}\sum_n \left[r_u^2 + r_v^2 + r_{\mathrm{div}}^2 + r_x^2 + r_y^2\right]_n \tag{36}$$

All five residual components in Eq. (36) have unit weights, and their squared sum is averaged over the N sampled points. Thus, $W = \frac{I_{5N}}{N}$ in Eq. (4), with no additional scaling of individual residual components. Mapping coordinates to the Chebyshev interval does not normalize these residuals. Equal weights are a fixed experimental choice; changing them may affect both coefficient fitting and candidate selection. No systematic weight-sensitivity study was performed, so the reported results are conditional on this choice.

The held-out test set provides out-of-sample evaluation over independently sampled points in the same benchmark domain. It does not constitute a continuous-domain error bound.

## 4.3 Chebyshev realization

The numerical realization uses tensor-product Chebyshev functions. After mapping each computational coordinate to the interval required by the Chebyshev representation, [10]

$$\phi_{abc} = T_a(\tilde{x})T_b(\tilde{y})T_c(\tilde{t}), \qquad D = a + b + c \tag{37}$$

A complete three-variable expansion through total degree $D$ contains

$$N_D = \binom{D+3}{3} \tag{38}$$

$$N_2 = 10, \quad N_6 = 84, \quad N_{10} = 286, \quad N_{13} = 560, \quad N_{20} = 1771 \tag{39}$$

Separate active sets are maintained for the two velocity components and pressure:

$$\hat{u} = \sum_{i\in A_u} a_{u,i}\,\phi_i, \qquad \hat{v} = \sum_{i\in A_v} a_{v,i}\,\phi_i, \qquad \hat{p} = \sum_{i\in A_p} a_{p,i}\,\phi_i \tag{40}$$

All derivatives required by the Navier-Stokes residual and candidate-response vectors are evaluated analytically from the Chebyshev representation.

## 4.4 PI-GMDH package variants

Adaptive package. Candidate Chebyshev functions are evaluated using $q$ and $m$. The implementation begins from a complete low-degree representation, searches the lowest unfinished shell and can continue into higher shells when required, retains candidates satisfying the configured $q$ threshold, prioritizes eligible candidates by $m$, and introduces them as a bounded package before coefficient reoptimization. The reported runs use $q \geq 0.05$ and a 256-term cap. The package-mass target is fixed after the initial degree-2 relaxation as 25% of the total eligible degree-3 projection mass, with a lower bound of $10^{-14}$. Eligible candidates are ordered by mean residual-interaction magnitude and accumulated until this target is reached or the 256-term cap is met; if necessary, the search continues into higher total-degree shells.

Complete degree-by-degree package. Every function in the next total-degree shell is activated before reoptimization. After completion of degree D, the representation contains all Chebyshev functions satisfying a+b+c ≤ D. In the main full-cell validation/test comparison, this policy was intentionally limited to Dmax = 13 to reduce computational cost because its role was to show the tendency of complete-shell growth rather than to determine its limiting accuracy at D = 20. A separate large-sample package-granularity experiment follows complete degree-by-degree growth through D = 20.

One-term package. One functional component is introduced per structural extension, followed by coefficient reoptimization. This provides a fine-grained limiting case at the cost of many optimization cycles.

All-terms package. The revised all-terms realization begins with the complete degree-2 representation containing 10 functions per field. After initial coefficient optimization, all remaining functions through total degree 20 are introduced in a single structural extension, producing 1,771 functions per field and 5,313 coefficients across $u$, $v$, and $p$, followed by further coefficient optimization.

These package variants are an ablation of structural construction, not different physical models. Adaptive and nonadaptive variants use the same candidate family and physical objective; what changes is how the next group of active functional components is formed.

## 4.5 Coefficient optimization and implementation

After a structural extension, coefficients are updated by alternating block-coordinate damped Gauss-Newton optimization. In the retained implementation, one relaxation cycle uses the sequence $u \to p \to v \to p$; pressure is updated twice because each velocity update changes the coupled momentum residual. The reported runs use $\lambda = 10^{-10}$, CuPy float64 arrays, 20,000-point chunks, GPU matrix products, and dense linear solves.

Residual and Jacobian contributions are accumulated without storing the complete Jacobian simultaneously. Coefficient optimization requires $J^T W J$ and $J^T W \mathbf{r}$, whereas adaptive candidate evaluation requires $\mathbf{j}_i^T W \mathbf{r}$ and $\mathbf{j}_i^T W \mathbf{j}_i$.

Because damped normal equations are used, conditioning of the active Jacobian remains a relevant numerical consideration, particularly for large nonselective representations.

### 4.6 External reference models

Selected physics-informed MLP and KAN configurations are included as external references. Several relatively simple configurations were examined and the best-performing configurations among those investigated were retained; no exhaustive architecture or hyperparameter search was performed. PI-GMDH structural-control parameters were examined more carefully during development of the present method than the neural-reference settings. The PINN and KAN results should therefore be interpreted as illustrative reference realizations rather than estimates of the best attainable accuracy or computational time of those model classes. Reported run times characterize the retained runs and do not include an unmeasured total cost of exploratory hyperparameter tuning. PI-GMDH variants likewise contain structural-control parameters, so their timings are timings of the specified runs rather than universal optimization costs. [3,8,9]

## 5. Results

### 5.1 Full-cell validation trajectories

Figure 1 shows the validation RMSE trajectories of the two observed velocity components. Adaptive PI-GMDH continues reducing both velocity errors over many orders of magnitude, whereas the reference models and the nonadaptive PI-GMDH variants level off at substantially higher values over the recorded trajectories. Figure 2 shows the corresponding pressure-gradient errors. Pressure-gradient values were not supplied as observational targets; their reconstruction is induced by the coupled Navier-Stokes momentum residuals.

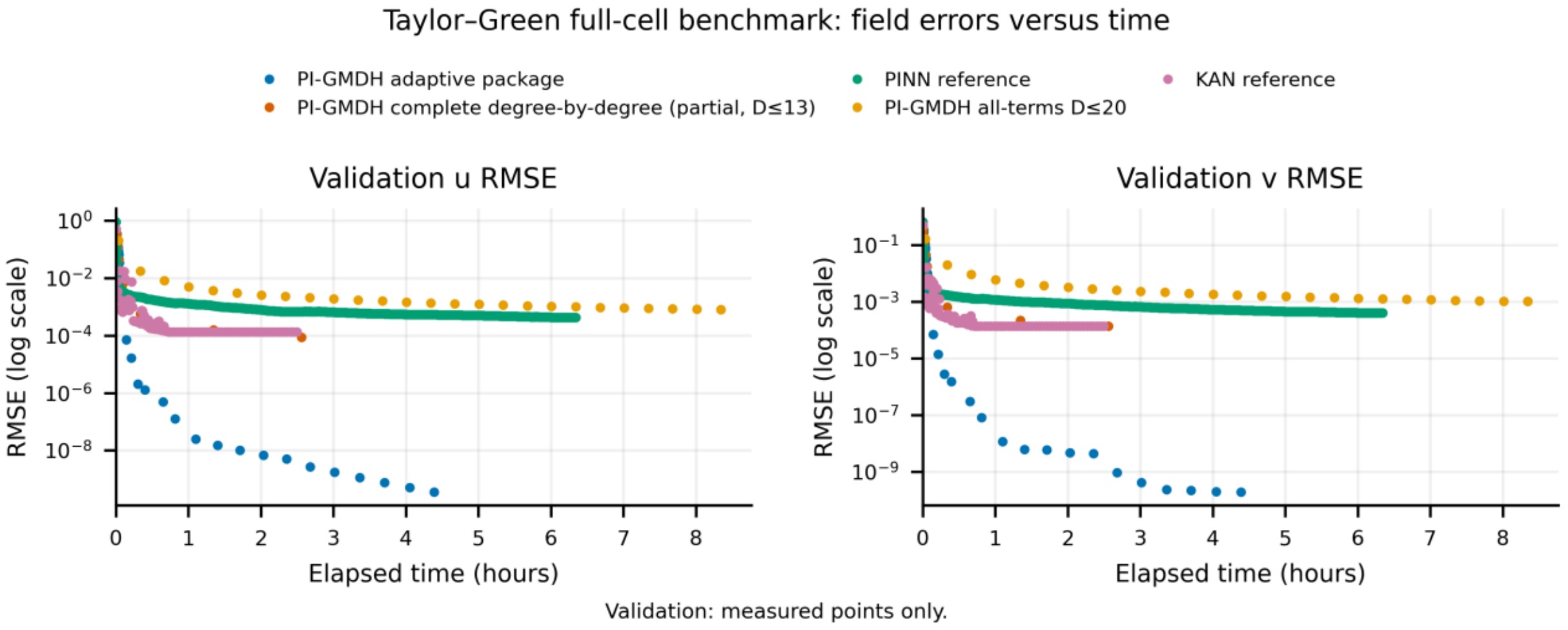

Figure 1. Validation velocity-component errors versus elapsed wall-clock time for the Taylor-Green full-cell benchmark. Points are recorded validation measurements without interpolation. The complete degree-by-degree trajectory was intentionally limited to Dmax = 13 to illustrate complete-shell growth at reduced computational cost.

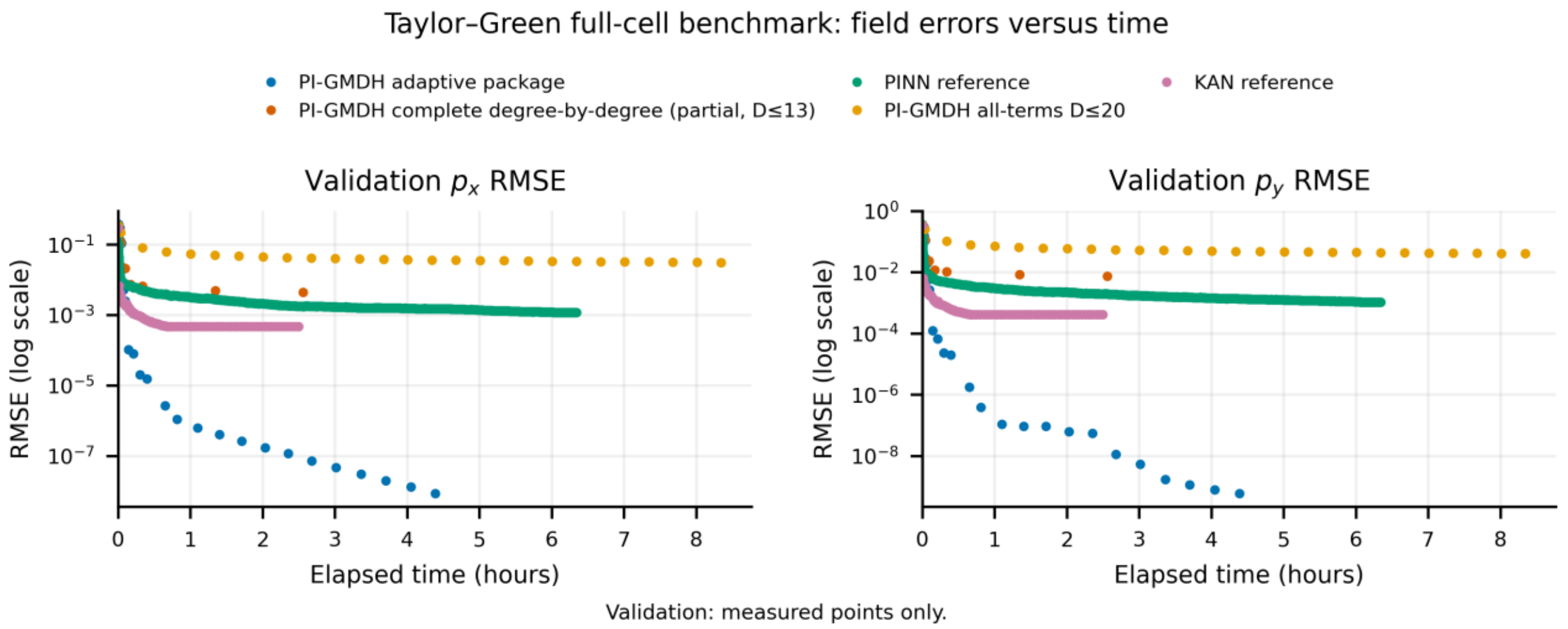

Figure 2. Validation pressure-gradient errors versus elapsed wall-clock time for the Taylor-Green full-cell benchmark. Pressure gradients were not supplied as observational targets; their reconstruction is induced through the coupled momentum equations. Unlike pressure itself, px and py are independent of the arbitrary pressure gauge. The complete degree-by-degree trajectory was intentionally limited to Dmax = 13.

The all-terms comparison is particularly informative because the adaptive and all-terms variants use the same underlying Chebyshev family. Their difference is therefore not simply representational capacity or availability of high-degree functions, but how the active representation is constructed and reoptimized during the solution process. The validation field-error trajectories show the same qualitative separation as the objective values reported below.

The complete degree-by-degree trajectory provides a second control. It retains progressive structural growth but activates complete degree shells rather than selectively constructed packages. In the main full-cell benchmark it is deliberately capped at Dmax = 13, so it is used to show the tendency of complete-shell growth and is not interpreted as the best attainable accuracy of that policy at larger maximum degree.

The separate package-granularity experiment in Section 5.2 complements this deliberately shortened full-cell trajectory by following complete degree-by-degree growth through D = 20.

## 5.2 Package granularity in a large-sample structural-growth experiment

A separate large-sample Taylor-Green experiment on $x,y \in [-1,1]$ and $t \in [0,1]$ was used to examine the computational effect of package granularity more directly. Approximately $10^6$ sampled points were used, making each coefficient-relaxation stage substantially more expensive than in the main validation/test protocol. Figure 3 compares the two limiting progressive-growth policies: activation of one functional component per structural extension and activation of a complete total-degree shell before reoptimization.

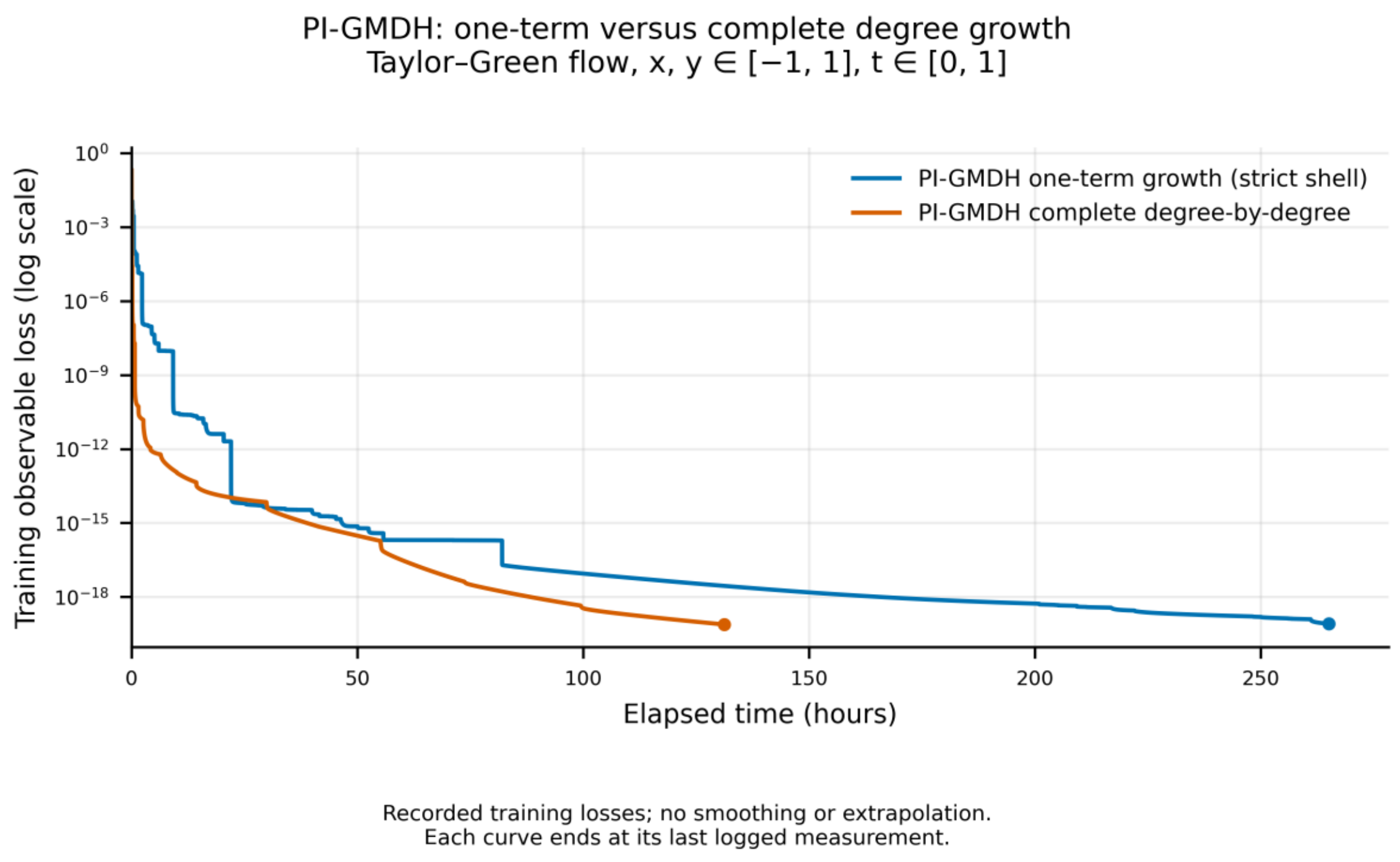

Figure 3. Effect of package granularity on PI-GMDH structural growth in the large-sample Taylor-Green experiment on x,y ∈ [−1,1], t ∈ [0,1]. The one-term trajectory was interrupted after reaching an incomplete total-degree-13 representation containing 555, 554, and 554 terms for u, v, and p, respectively; a complete degree-13 basis contains 560 terms per field. The complete degree-by-degree trajectory continued through total degree 20. The one-term endpoint therefore represents an interrupted structural-growth trajectory and is not a completed degree-20 calculation.

The trajectories expose a trade-off between structural selectivity and optimization overhead. The one-term policy reached the approximately $10^{-19}$ loss regime while using an almost complete degree-13 representation, but required repeated coefficient reoptimization after individual structural additions and was still running when interrupted. Complete degree-by-degree growth amortized the relaxation cost over whole shells and reached degree 20 in substantially less wall-clock time. The comparison motivates intermediate package construction: packages can preserve selective structural growth while reducing the number of expensive reoptimization stages relative to one-term activation.

## 5.3 Validation and held-out test evaluation

The validation-selected checkpoints were evaluated on the independently sampled test set. The results are summarized in Table 1. Adaptive PI-GMDH attained a validation loss of $5.299\times10^{-19}$ and a test loss of $5.296\times10^{-19}$. The close agreement between these values shows that the very small sampled objective obtained during construction is reproduced on independently sampled points from the same benchmark domain rather than being restricted to the training sample.

Table 1. Validation-selected benchmark results. *The complete degree-by-degree run was intentionally limited to Dmax = 13 to reduce computational cost and to show the tendency of complete-shell growth; it is not intended to establish the limiting D = 20 performance of that policy. †Neural reference models do not have field-wise Chebyshev active-function counts.

| Method | L_val | L_test | RMSE u | RMSE v | RMSE p_x | RMSE p_y | Active functions u/v/p | Time (h) |
|---|---|---|---|---|---|---|---|---|
| PI-GMDH adaptive package | 5.299e−19 | 5.296e−19 | 3.567e−10 | 1.948e−10 | 8.508e−9 | 5.891e−10 | 204 / 201 / 175 | 4.40 |
| PI-GMDH complete degree-by-degree (Dmax=13)* | 5.873e−8 | 5.776e−8 | 8.559e−5 | 1.378e−4 | 4.310e−3 | 7.413e−3 | 560 / 560 / 560 | 2.56 |
| PI-GMDH all-terms (Dmax=20) | 2.776e−6 | 2.722e−6 | 7.992e−4 | 1.018e−3 | 3.071e−2 | 4.064e−2 | 1771 / 1771 / 1771 | 8.35 |
| PINN reference | 1.754e−6 | 1.739e−6 | 4.232e−4 | 3.999e−4 | 1.178e−3 | 1.062e−3 | N/A† | 6.34 |
| KAN reference | 4.127e−7 | 4.132e−7 | 1.342e−4 | 1.385e−4 | 4.697e−4 | 4.173e−4 | N/A† | 2.50 |

The complete degree-by-degree PI-GMDH realization, intentionally limited to Dmax = 13, used the complete 560-function basis in each field and produced a validation loss of $5.873\times10^{-8}$ and a held-out test loss of $5.776\times10^{-8}$ in 2.56 h. Its held-out RMSEs were $8.559\times10^{-5}$ for u, $1.378\times10^{-4}$ for v, $4.310\times10^{-3}$ for px, and $7.413\times10^{-3}$ for py. Because the purpose of this run is to display the tendency of nonselective complete-shell growth at reduced cost, it should not be interpreted as the limiting accuracy of complete degree-by-degree PI-GMDH at D = 20.

The selected PINN and KAN references produced test losses of $1.739\times10^{-6}$ and $4.132\times10^{-7}$, respectively. Their velocity RMSEs were in the $10^{-4}$ range, while pressure-gradient RMSEs ranged from $4.173\times10^{-4}$ to $1.178\times10^{-3}$. These results are reference realizations rather than estimates of the best attainable performance of the neural model classes because no exhaustive architecture or hyperparameter search was performed. The complete degree-by-degree realization has a lower total test objective than these selected neural references at Dmax = 13, although its individual field RMSEs are not uniformly lower; the total objective also contains divergence and momentum residual contributions. These numerical comparisons describe only the retained configurations and are not evidence that the reference approaches cannot achieve comparable or better results under more extensive tuning.

The structural ablation against the all-terms PI-GMDH variant is particularly important. The all-terms model activated 1771 functions for each field, or 5313 coefficients overall, yet produced a test loss of $2.722\times10^{-6}$. Adaptive PI-GMDH used 580 active functions overall, approximately 9.2 times fewer, required 4.40 h rather than 8.35 h, and reduced the held-out objective by more than twelve orders of magnitude relative to the all-terms realization. Because both variants use the same Chebyshev candidate family and physics-informed objective, the comparison demonstrates the sensitivity of the attained results to the combined structural-construction and coefficient-optimization trajectory. It does not separately identify which component of the adaptive integration causes the observed improvement.

The lower loss attained by the adaptive realization does not contradict the inclusion of its active functions in the complete degree-20 basis. Setting the additional coefficients to zero allows the full representation to reproduce the adaptive solution; consequently, its global minimum training objective cannot be higher when the objective and training samples are identical. However, the reported results are held-out test losses of validation-selected checkpoints obtained through finite nonlinear optimization, rather than globally minimized linear least-squares objectives. Although each Gauss-Newton update solves a linearized coefficient problem, the coupled Navier-Stokes residual remains nonlinear in the field coefficients. Progressive activation changes both the intermediate system sizes and the optimization trajectory, so the two procedures need not reach the same solution under their stopping criteria. Smaller intermediate systems can reduce the cost of individual coefficient updates, but the present experiments do not establish whether system size, numerical conditioning, or other aspects of the optimization trajectory primarily explain the observed accuracy difference.

The adaptive representation remained comparatively compact, containing 204, 201, and 175 active functions for u, v, and p, respectively. The corresponding held-out test RMSEs were $3.567\times10^{-10}$ and $1.948\times10^{-10}$ for the two velocity components. Pressure and pressure gradients were not supplied as direct training targets; nevertheless, the recovered pressure gradients reached RMSEs of $8.508\times10^{-9}$ for px and $5.891\times10^{-10}$ for py.

### 5.4 Representation size and structural efficiency

The all-terms degree-20 realization contains 1,771 functions for each of u, v, and p, or 5,313 coefficients overall. The complete degree-by-degree Dmax = 13 realization contains 560 functions per field, or 1,680 overall. By contrast, the validation-selected adaptive representation contains 204, 201, and 175 active functions, respectively, for a total of 580. The adaptive representation therefore uses approximately 10.9% of the coefficient count of the all-terms representation and 34.5% of the coefficient count of the complete degree-13 representation. This difference is structural rather than merely a consequence of a lower search degree: reaching a given total degree during adaptive exploration does not imply that all functions through that degree have been activated.

The combination of active-representation size, elapsed time, and held-out error provides a direct view of structural efficiency. In the present benchmark, the adaptive package policy achieves the smallest validation and test objectives while using far fewer active functions than either nonselective Chebyshev construction. The intentionally capped complete degree-by-degree result additionally shows that progressive complete-shell growth can outperform the one-shot all-terms representation even with a substantially smaller basis, emphasizing that the construction path matters in addition to final nominal capacity.

## 6. Discussion

The proposed method sits at the intersection of several established ideas rather than replacing them. From GMDH it inherits progressive structural self-organization [1,2]; from greedy function-space approximation it shares the idea of evaluating directions in a functional representation [15]; and from adaptive PDE methods it shares residual-driven enrichment of an approximation space [14]. In the proposed adaptive implementation, candidate directions are scored after propagation through the complete coupled physics-informed residual and are activated in packages before the active coefficient blocks are reoptimized.

The experiments are designed primarily to investigate structural construction of a physics-informed functional representation. The comparison among PI-GMDH variants is consequently more informative for this purpose than a direct ranking against neural architectures. Adaptive, complete-degree, one-term, and all-terms variants can share the same physical residual, functional family, and coefficient optimizer while differing in how packages are constructed. This makes package policy an explicit experimental variable.

The large-sample package-granularity experiment in Figure 3 makes the computational trade-off explicit. One-term growth had accumulated 1,663 active functions (555/554/554 for u/v/p) when it was interrupted within degree 13, whereas complete degree-by-degree growth continued through degree 20. The one-term trajectory nevertheless reached the same order of training loss shown by the completed degree-growth trajectory, but at much greater wall-clock cost. Because the one-term run was interrupted and the experiment used a different, approximately million-point sampling protocol on $x,y \in [-1,1]$, these timings are used only to illustrate structural-growth behavior and are not merged with the full-cell validation/test comparison in Table 1.

The validation and test results strengthen the inference suggested by the structural trajectories: making a large functional space available is not sufficient by itself. The all-terms degree-20 model has access to the complete Chebyshev space through the prescribed maximum degree and contains 5,313 coefficients, yet its held-out test loss is $2.722\times10^{-6}$. Adaptive PI-GMDH uses only 580 active functions and reaches $5.296\times10^{-19}$ on the same test protocol. The deliberately limited complete degree-by-degree $D_{max} = 13$ realization, with 1,680 coefficients, reaches $5.776\times10^{-8}$, demonstrating that the path of progressive shell-wise construction can itself change the attainable result relative to one-shot activation of a larger space. Since these PI-GMDH realizations share the candidate family and physics-informed formulation, the observed separation is associated with construction and optimization path rather than with Chebyshev approximation alone. The comparison is also not determined only by the ultimate accuracy that a construction policy might eventually attain. Even if continued complete degree-by-degree growth or additional optimization of the all-terms representation were able to approach the same accuracy, the wall-clock time and number of reoptimization stages required to reach a given error level would remain practically important. Accordingly, the present timings are interpreted as part of an accuracy–cost trade-off under the tested implementation and hardware, rather than as evidence that the alternative construction policies cannot ultimately reach comparable accuracy. The present comparison examines package-construction policy (selective versus non-selective, and fine versus coarse structural growth) but does not separately isolate the contribution of evaluating candidate directions against the complete coupled physical residual. In particular, a generic greedy selector based only on observational residuals has not been examined under the same construction protocol. The present results are therefore consistent with, but do not independently establish, the specific advantage of physics-guided candidate scoring over generic greedy structural selection. Isolating this contribution is an important subject for further investigation. This interpretation should not be overstated as a demonstrated conditioning result. In a discrete coupled residual system, mathematically independent Chebyshev functions can nevertheless produce residual-response columns with unfavorable numerical dependence, and normal equations can amplify conditioning difficulties. Condition numbers or singular-value spectra should be measured before conditioning is claimed as the mechanism responsible for the observed plateaus.

Pressure recovery provides an additional test of the coupled physical construction. Pressure and its gradients are absent from the direct observational targets, so their reconstruction is driven through the momentum residuals. The adaptive model attains pressure-gradient RMSEs of $8.508\times10^{-9}$ for px and $5.891\times10^{-10}$ for py on the held-out set. This does not remove the pressure-gauge ambiguity, but it demonstrates that the gradient information required by the governing equations can be recovered to high accuracy in this benchmark through coupling to the observed velocity fields.

The adaptive package policy used here is itself only one realization of PI-GMDH. Its threshold, package size, hierarchy exploration, and balance between structural extension and coefficient relaxation are control parameters.

Several of these decisions are associated with quantities observable during the evolving solution, including distributions of $q$ and $m$, current objective reduction, active support size, and effectiveness of recent packages. This suggests a future direction in which package construction is controlled online rather than by fixed rules. Contextual-bandit, reinforcement-learning, or simpler adaptive controllers could adjust candidate thresholds, package size, search depth, or the balance between structural growth and coefficient optimization.

The PINN and KAN comparisons describe the retained configurations and their recorded training trajectories. These reference models were not exhaustively tuned, while PI-GMDH structural-control parameters received more exploration during development. Table 1 does not establish a general accuracy-time advantage: the selected KAN run has a shorter runtime but a substantially larger error, whereas adaptive PI-GMDH reaches a lower error than both neural references and finishes sooner than the selected PINN run. Figures 1 and 2 show how the field errors evolve with elapsed time under these particular settings. They do not establish the best attainable performance of any model class or exclude comparable or better neural results with additional training or tuning. Broader comparisons would require systematic parameter searches and matched computational budgets. The present results motivate further investigation of adaptive structural construction rather than a general ranking of PI-GMDH, PINNs, and KANs. [3,8,9]

The numerical evidence is restricted to reconstruction of a single smooth, analytical, two-dimensional Taylor-Green flow using dense, noise-free velocity observations. Tensor-product Chebyshev functions are particularly well suited to this smooth benchmark. The reported accuracy and computational behavior therefore characterize the tested combination of problem, functional family, sampling protocol, and optimization settings; they do not establish comparable performance for other differential systems, nonsmooth solutions, sparse or noisy observations, or different flow regimes. Although pressure and its gradients are not supplied as observational targets, their recovery is constrained through the momentum equations by the observed velocity fields. This experiment consequently evaluates physics-informed field reconstruction and does not demonstrate a forward Navier-Stokes solution driven primarily by initial and boundary conditions. Independent validation and test samples assess performance within the same benchmark domain, rather than transfer to other problems. Additional differential systems, sparse and noisy observations, and initial-boundary-value problems remain necessary subjects for future evaluation.

## 7. Conclusions

This work formulates a Physics-Informed Method of Group Data Handling in which the structure of a functional approximation is constructed progressively during solution. PI-GMDH separates structural construction from parameter optimization: packages of functional components extend the active representation, after which their coefficients are optimized against the complete physical and observational objective.

Within this framework, a first-variation-based adaptive package strategy uses the governing equations to evaluate inactive functional directions before they are introduced. Normalized residual alignment measures directional relevance, while mean residual-interaction magnitude provides complementary information about the strength of the candidate interaction with the current residual. The same residual-response vectors subsequently form Jacobian columns used for coefficient optimization.

The Taylor-Green study is organized as an ablation of package construction. Adaptive physics-guided packages, complete degree-by-degree packages, one-term packages, and a one-shot all-terms package use a common Chebyshev family and common physics-informed formulation but expose the coefficient optimizer to different evolving active spaces. Adaptive PI-GMDH reached a validation loss of $5.299\times10^{-19}$ and a held-out test loss of $5.296\times10^{-19}$ with 204/201/175 active functions for u/v/p. The intentionally limited complete degree-by-degree Dmax = 13 realization reached a test loss of $5.776\times10^{-8}$ with 560 functions per field, while the all-terms degree-20 realization used 1771 functions per field but produced a test loss of $2.722\times10^{-6}$. For this smooth Taylor-Green reconstruction problem with dense, noise-free velocity observations, the results demonstrate that package-construction policy can substantially affect the attained accuracy and computational cost within the tested Chebyshev representation and optimization procedure. The close agreement between adaptive validation and test losses indicates that the very small sampled residual is reproduced on independently drawn points from the same domain. It should not, however, be interpreted as a continuous-domain error bound, as proof that physics-guided scoring alone causes the observed improvement, or as evidence of universal superiority over other model classes. The neural comparisons establish only the errors and elapsed times observed for the tested configurations. They do not demonstrate a general accuracy-time advantage or determine what PINNs and KANs could attain with further training and systematic tuning. Future work should examine data-only versus full-physics structural selection, additional differential systems, sparse and noisy data, initial-boundary-value formulations, more robust linear solvers, and online adaptation of the package-construction policy.

## Funding

This research was partially supported by the U.S. Department of State under Grant No. NSEPEUR23002-1019.

## Declaration of Generative AI and AI-Assisted Technologies

OpenAI Codex was extensively used during the development, debugging, and refinement of the research software. OpenAI GPT was extensively used during preparation of the manuscript, including assistance with scientific writing, restructuring, language editing, and refinement of the mathematical presentation. The research methodology, selection and interpretation of numerical experiments, assessment of the results, and scientific conclusions were determined by the author. All AI-assisted software and manuscript content were reviewed by the author, who takes responsibility for the final content of the work.

## Data and Code Availability

The source code used to implement PI-GMDH and reproduce the numerical experiments presented in this work is publicly available in the PIMGDH repository on GitHub: https://github.com/mininmy/PIMGDH. The benchmark definitions and experimental settings required to reproduce the reported calculations are described in the manuscript and repository.